\documentclass[a4paper,twoside]{article}

\usepackage{url}
\usepackage{epsfig}
\usepackage{subcaption}
\usepackage{calc}
\usepackage{amssymb}
\usepackage{amstext}
\usepackage{amsmath}
\usepackage{amsthm}
\usepackage{stackrel}
\usepackage{multicol}
\usepackage{pslatex}
\usepackage{apalike}
\usepackage{color}
\usepackage{listings}
\usepackage[linesnumbered]{algorithm2e}
\usepackage{SCITEPRESS}     

\newcommand{\Aplib}{{\sf Aplib}}
\newcommand{\aplib}{{\sf aplib}}

\newcommand{\HIDE}[1]{}
\newcommand{\xblind}[1]{#1}

\newtheorem{mydef}{Def.}

\begin{document}
\urlstyle{same}

\title{Aplib: Tactical Programming of Intelligent Agents}

\author{\authorname{\xblind{I. S. W. B. Prasetya\sup{1}\orcidAuthor{0000−0002−3421−4635}}}
    \affiliation{\sup{1}\xblind{Utrecht University, the Netherlands}}
    \email{\xblind{s.w.b.prasetya@uu.nl}}
}


\abstract{This paper presents \aplib, a Java library for programming intelligent agents,
featuring BDI and multi agency, but adding on top of it a novel layer of
tactical programming inspired by the domain of theorem proving.
\Aplib\ is also implemented in
such a way to provide the fluency of a Domain Specific Language (DSL).
Compared to dedicated BDI agent programming languages such as JASON, 2APL, or GOAL,
\aplib's embedded DSL approach does mean that \aplib\ programmers will still
be limited by Java syntax, but on other hand they
get all the advantages that Java programmers get:
rich language features (object orientation,
static type checking, $\lambda$-expression, libraries, etc),
a whole array of development tools, integration with other technologies,
large community, etc.}

\onecolumn \maketitle \normalsize \setcounter{footnote}{0} \vfill


\section{\uppercase{Introduction}}
\label{sec:introduction}

\vspace{-2mm}
Software agents are generally considered as a different type of programs
than e.g. procedures and objects as they are naturally autonomous,
reactive as well as pro-active, and 'social' (they interact with each other)
\cite{wooldridge1995intelligent,MeyerAgentTech2008}. As such, they are considered
as highly suitable building blocks to build complex software systems that require
multiple and decentralized loci of control \cite{jennings2001agent}.
Applications of software agents include computer games, health care,
traffic control system \cite{jennings1998agent},
smart electrical power grid control \cite{merabet2014applications},
and manufacturing control system \cite{leitao2009agent}.

In a stronger concept of agency \cite{wooldridge1995intelligent}, agents
can also posses artificial intelligence. The most popular type of intelligent agents is probably
that of the BDI (Belief-Desire-Intent) family \cite{herzig2017bdi}. Such an agent maintains a set of
human-inspired mental states, such as belief, desire, and intention, and is able
to reason over these states when deciding its actions.
While adding AI would greatly enhance agents, it is not something
that we get for free as the AI would need some programming in the first place.
In the case of BDI agents, someone would need to produce the set of
inference rules that control their actions. There are indeed
programming languages to program BDI agents, e.g.
   JASON \cite{bordini2007programming}, 
   2APL \cite{dastani20082apl},
   GOAL \cite{GOALprogmanual2018},
   JACK \cite{winikoff2005jack}, 
   FATIMA \cite{dias2014fatima},
   and PROFETA \cite{fichera2017python},
that allow the rules to be declaratively formulated, but this does not necessarily mean
that it is easy for agent programmers to develop these rules, especially
if the problem to solve is complex.

This paper presents \aplib\footnote{\xblind{\url{https://iv4xr-project.github.io/aplib/}}}: a BDI agent programming framework
that adds as novelty a layer of {\em tactical programming} over the rule based
programming typical in BDI agent programming.
Tactics allow agents to strategically choose and prioritize their short term plans.
This is inspired by proof programming in LCF theorem provers like Coq and HOL
\cite{delahaye2000tactic,gordon1993introduction}. These theorem provers come with a whole range of proof rules.
However, having plenty of rules does not in itself make proving formulas
easy. In fact, proving a complex goal formula often involves interactively
trying out different steps, searching for the right sequence that would solve the goal.
To help  users, these theorem provers provide tactical combinators to compose
tactics from proof rules, hence users can write proofs by sequencing
tactics rather than proof rules.
There is some analogy with agents, which also have to solve non-trivial goals,
hence inspiring \aplib\ to provide a similar tactical approach to program
agents.

While tactics are good to capture bottom-up strategies to solve a goal\footnote{With respect
to the previously mentioned concept of 'tactic' in LCF theorem provers, \aplib\ tactics
express bottom-up strategies, whereas LCF tactics are top-down. Aside from the directions,
both concepts intend to express strategical composition of the underlying basic steps.}, sometimes it is also
useful to have top-down strategies. This can be expressed by a way to break down a goal
into subgoals. \Aplib\ facilitates this through the concept of {\em goal structure}, that allows
a goal to be hierarchically formulated from subgoals through a number of strategy combinators,
e.g. to express fall backs (alternate goals if the original goal fails).
%
%
While it is true that
tactics and goal structures can be programmed inside BDI agents' reasoning rules, we would argue
that tactical programming involves a different mental process for programmers. \Aplib\ allows
them to be programmed separately and more abstractly, rather than forcing the programers to encode them inside
reasoning rules.


Unlike JASON, 2APL, or GOAL, which offer a native/dedicated BDI agent programming,
\aplib\ offers a Domain Specific Language (DSL) to program agents,
{\em embedded} in Java. This means that \aplib\ programmers will program in Java,
but they will get a set of APIs that give the fluent appearance of a DSL.
In principle, having a native programming language is
a huge benefit, but {\bf only} if the language is mature and scalable.
On the other hand, using an embedded DSL means that the programmers have direct access
to all the benefit the host language, in this case Java:
its expressiveness (OO, lambda expression etc), static typing,
rich libraries, and wealth of development tools.
These are things that JASON, 2APL, nor GOAL cannot offer. It is also worth noting that elsewhere,
the rapidly growing popularity
of AI and data science {\em libraries} like TensorFlow, NumPy, and Java-ML can be seen as
evidence that developers are quite willing to sacrifice the convenience of having a
native language in exchange for strength.

{\bf Paper structure.} Section \ref{sec.notation} first introduces some notation and concepts
from OO programming that might be non-standard.
Related work will be discussed later, namely in Section \ref{sec.relatedwork}.
Section \ref{sec.aplibagents} explains the basic concepts of \aplib\ agents and
shows examples of how to create an agent with \aplib\ and how to write some
simple tactics. The section also presents \aplib's deliberation algorithm, which necessarily
has to extend the standard BDI deliberation algorithm.
Section \ref{sec.tactic} presents \aplib's tactical programming.
Section \ref{sec.reasoning} discusses \aplib's reasoning backend,
and Section \ref{sec.structuredgoal} presents \aplib's goal structuring mechanism and
also discusses budgeting as a means to control agent's commitment to goals.
Finally Section \ref{sec.concl} concludes and mentions some future work.

\vspace{-3mm}
\section{Preliminary} \label{sec.notation}

\vspace{-3mm}
{\bf Notation.}
Since \aplib\ is implemented in Java, most of its concepts are implemented as objects.
Objects can be structured {\em hierarchically}. We use the notation $u_{{\mapsto} v}$ to denote that the object
$v$ is linked from $u$ through a reference inside $u$ (and therefore its state is also reachable
from $u$). This is useful for abstraction, as sometimes it is convenient to talk about
the structure $u_{{\mapsto} v}$ in terms of its parent $u$ while being implicit about the
subobject $v$.

{\bf Java functions.}
Since Java 8, functions can be conveniently formulated using so-called
lambda expressions. E.g. the Java expression:
\vspace{-1mm}
\[ x \rightarrow  x{+}1 \]
%
constructs a nameless function that takes one parameter, $x$, and returns the value of $x{+}1$. \Aplib\
relies heavily on functions. Traditionally, to add behavior to an object we do that
by defining a method, say $m()$, in the class $C$ to which the object belongs to. But then all
objects of $C$ will have the same behavior. If we want to assign a different behavior to
some of them we have to first create a subclass $C'$ of $C$ where we override $m()$ with the new
behavior, and then instantiate $C'$ to obtain new objects with the new behavior.
If we expect to do this often, this approach will clutter the code.
Instead, in \aplib\ we often apply a design pattern similar to the Strategy Pattern \cite{designpatterns94}, where
we implement $m$ as a field of type $\sf Function$. If we have an object $u:C$,
and we want to change the behavior of its $m$, we can simply assign
a new function to it, as in $u.m = x {\rightarrow} x{+}1$. There is no overhead of having
to create a subclass.

Unlike in a pure functional language like Haskell, Java functions can be either
{\em pure} (has no side effect) or impure/effectful.
An {\em effectful} function of type $C {\rightarrow} D$ takes
an object $u : C$ and returns some object $v:D$, and may also alter the state of $u$.

\vspace{-3mm}
\section{\Aplib\ Agency} \label{sec.aplibagents}

\begin{figure}
    \begin{center}
    \includegraphics[scale=0.3]{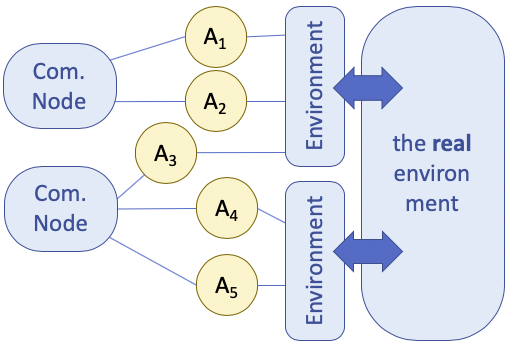}
    \end{center}
    \caption{\em Typical deployment of \aplib\ agents. In the picture,
    $A_i$ are agents. "Com. nodes" allow connected agents to send messages to each other.
    In \aplib\ terminology, an 'environment' is an interface to the real environment. The agents themselves will not see the distinction.
    }
    \label{fig.deploy}
\end{figure}

\vspace{-2mm}
Figure \ref{fig.deploy} illustrates the typical way \aplib\ agents are deployed.
As common with software agents, \aplib\ agents are intended to be used in conjunction with
an {\em environment} (the 'real environment' in Fig. \ref{fig.deploy}), which is assumed to run autonomously for its own purpose.
This environment can be e.g. a computer game, a simulator, a trading system, or
a manufacturing system as in \cite{leitao2009agent}. Each agent would have
its own goal, which can be to simply monitor the environment, or to influence
it in a certain way. Some agents may work together towards a collective goal,
whereas others might be competitors.
A group of agents that wish to collaborate can register to a 'communication node'
(see Fig. \ref{fig.deploy}). Agents sharing the same node can send messages
to each other (singlecast, broadcast, or role-based multicast).

Agents are assumed to have {\em no} direct access to the real environment's state, e.g. due to security concerns.
Instead, an agent can 'sense' the environment to
obtain insight on a part of its state that the agent
is allowed to see. To influence the environment, the agent can send commands
to the latter, from a set of available commands. The environment can also sends messages
to the agent, e.g. to notify it that something that might interest the agent
just happened. The same facility can also be used by agents to send messages to
each others. Sending messages to \aplib\ agents is an asynchronous operation, hence
the environment is not slowed (or worse: locked) when it tries to send a message
to an agent. Sending a command to the environment is a synchronous operation:
the sending agent halts until it gets a return value from the environment
confirming if the command succeeded or failed.

{\bf BDI with goal structure.}
As typical in BDI (Belief-Desire-Intent) agency, an \aplib\ agent has a concept of
belief, desire, and intent. Its belief is simply the information it has in its own state,
which includes information on what it believes to be the current state of the real environment
(due to the asynchronous nature of the above described agents-environment system, this
is not necessarily the same as the environment's actual state). The agent can be
given a {\em goal structure}, defining the its desire. Unlike flat structured
goal-base used e.g. in 2APL and GOAL, a goal structure is richly structured,
with different nodes expressing different ways of how a goal could be achieved
through its subgoals. More on this will be discussed Section \ref{sec.structuredgoal}.

Abstractly, an \aplib\ agent is a tuple:
\[ A \ = \ (s_{{\mapsto} E},\Pi,\beta) \]
where $s$ is an object representing $A$'s state and
$E$ is its environment. More precisely, $E$ is an interface abstracting the
real environment as depicted in Fig. \ref{fig.deploy}. $A$ does not have access to the real environment, though it can
see whatever information that the real environment chooses to mirror in $E$.
When $A$ wants to send a command to the environment, it actually sends it to $E$,
which under the hood will handle how it will be forwarded to the real environment.

$\Pi$ is a goal structure, e.g. it can be a set of goals that have to be
achieved sequentially. Each goal has the form $g_{{\mapsto}T}$,
where $T$ is a 'tactic' intended to solve it. When the agent decides to
work on a goal $g_{{\mapsto}T}$, it will commit to it. In BDI terms, this reflects intention:
it will apply $T$ {\em repeatedly} over multiple execution cycles until
$g$ is achieved/solved, or the agent has used up its 'budget' for $g$.

{\bf Budget.}
To control how long the agent should persist on pursuing its current goal,
the component $\beta$ specifies how much computing budget the agent has.
Executing a tactic consumes some budget. So, this is only possible if $\beta{>}0$.
Consequently, a goal will automatically fail when $\beta$ reaches 0.
Budget plays an important role when dealing with a goal structure with multiple
goals as the agent will have to decide how to divide the budget over different
goals. This will be discussed later in Section \ref{sec.structuredgoal}.

\begin{figure}
    \begin{center}
    \includegraphics[scale=0.25]{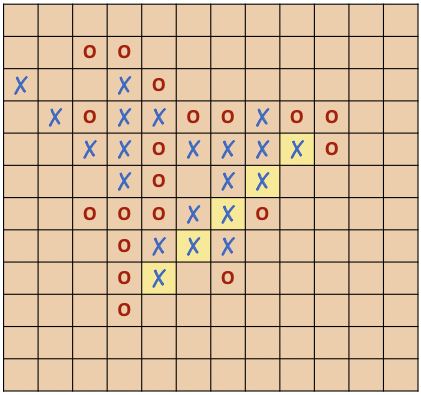}
    \end{center}
    \caption{\em A GoMoku game on a $12{\times}12$ board. Cross wins the game with a winning diagonal (yellow).}
    \label{fig.gomoku}
\end{figure}

{\bf Example.} As a running example suppose we want to develop an agent to
play a well known board game called {\em GoMoku}. The game is played on
a board consisting of $N{\times}N$ squares.
Two players take turn to put one piece every turn, a cross for player-1,
and a circle for the other. The player that manages to occupy five consecutive
squares, horizontally, vertically, or diagonally, with his own pieces, wins.
Fig. \ref{fig.gomoku} shows an example of a GoMoku board.

Figure \ref{fig.createagent} shows how we create this agent in \aplib. We
call it $\sf Crosy$ (it is intended to play the role of player-1, with cross).
Lines \ref{gomoEnvStart}-\ref{gomoEnvEnd} show the relevant
part of the environment the agent will use to interface with the actual GoMoku game.
It has e.g. the method ${\sf move}(t,x,y)$ to place a piece of type $t$ (cross or circle)
in the square $(x,y)$.

Lines \ref{line:createAgent}-\ref{line:setgoal} creates the agent.
It shows that a fresh state is created and attached
to the agent (line \ref{line:createstate}). Assuming $\sf gomEnv$ is an initialized instance
of $\sf GoMokuEnv$ (defined in lines \ref{gomoEnvStart} - \ref{gomoEnvEnd}),
line \ref{line:attachenv} hooks this environment to the agent.
Line \ref{line:setgoal} assigns the goal $\Pi$ to the agent,
defined in lines \ref{line:goalstart}-\ref{line:goalend},
stating that the desired situation is where the game is won by cross
(line \ref{line:goal}). Line \ref{line:goalend} associates the tactic $T$ (its definition is not shown)
to this goal, which the agent will use to solve the latter.
%

\begin{figure}
\begin{lstlisting}[mathescape=true,
    numbers=right,
    numberstyle=\tiny,
    numbersep=0pt
    ]
class GoMokuEnv extends Environment {                        $\label{gomoEnvStart}$
   static String CROSS  = "cross" ;
   static String CIRCLE = "circle" ;
   void move(String ptype, int x, int y) ...
   boolean crossWin() ...
   Set<Square> emptySquares() ...
}                                                           $\label{gomoEnvEnd}$
var $\Pi$ = goal("g")                                          $\label{line:goalstart}$
 . toSolve(s $\rightarrow$ s.env.crossWin())                   $\label{line:goal}$
 . tactic($T$) ;                                               $\label{line:goalend}$

var Crosy = new BasicAgent()        $\label{line:createAgent}$
 . withState(new AgentState())  $\label{line:createstate}$
 . withEnvironment(gomEnv)      $\label{line:attachenv}$
 . setGoal($\Pi$) $\label{line:setgoal}$
 . budget(200)
\end{lstlisting}
\caption{\em Creating an agent named $\sf Crosy$ to play GoMoku. Note that the code above is in
Java. \Aplib\ is not a separate programming language. Instead, it is a DSL embedded in Java.
The notation $x{\rightarrow}e$ in line \ref{line:goal}
is Java lambda expression defining a function (see also Section \ref{sec.notation}), in this
case a predicate defining the goal.} \label{fig.createagent}
\end{figure}

\subsection{Action}\label{sec.action}

\vspace{-3mm}
A tactic is made of so-called actions, composed hierarchically to define a goal-solving strategy.
Such composition will be discussed in Section \ref{sec.tactic}.
In the simple case though, a tactic is made of just a single action.
An {\em action} is an effectful and guarded function over the
agent state. The example below shows the syntax for defining an action.
It defines an action with "id" as its id, and binds the action to the Java variable $\alpha$:

{\small
\begin{equation}  \label{ex.fluent.interf}
{\bf var} \; \alpha \; = {\bf action}("id").\; {\bf do\_}(f).\; {\bf on\_}(q)
\end{equation}}

Above\footnote{
Note that $\bf action$, $\bf do\_$, and $\bf on\_$ are not Java keywords.
They are just methods. However, they are written to also implement the
Fluent Interface Pattern \cite{fowler2005fluent}. It is a design pattern
commonly used in embedded Domain Specific Languages (DSLs) to
'trick' the syntax restriction of the host language to allow them to
called in a sequence as if they form a sentence to improve the
fluency of the DSL.
},
$f$ is a function
defining the behavior that will be invoked when the action $\alpha$ is executed.
This function is effectful and may change the agent state.

The other, $q$, is a pure function specifying the 'guard' for the action.
Rather than using a predicate as a guard, which would be natural, we will allow
$q$ to be written as a query. More precisely, let $\Sigma$ be the type of the agent state,
we allow $q$ to be a function of type $\Sigma{\rightarrow} R$. So, it can be inspected
on a state $s$, to return some value of type $R$.
We treat $q$ as a predicate: $\widehat{q}(r,s) \; \stackrel{d}{=} \; (q(s)=r) \wedge r\not={\bf null}$.
The action $\alpha$ is only executable if it is enabled; it is enabled on
a state $s'$ if $\widehat{q}(r,s')$ is satisfiable
(there is an $r$ that would make it true). The behavior function $f$
has the type $\Sigma{\rightarrow}R{\rightarrow V}$ for some type $V$.
When  the action $\alpha$ is executed on $s$, it invokes $f(s)(r)$, where
$r$ is the solving value of the predicate  $\widehat{q}(r,s)$\footnote{
   This scheme of using $r$ essentially simulates {\em unification} a la pgrules in 2APL.
   Unification plays an important role in 2APL.
   The action in (\ref{ex.fluent.interf}) corresponds to pgrule $\widehat{q}(r)? \; | \; f(r)$
   The parameter $s$ (the agent's state/belief) is kept implicit in pgrules.
   In 2APL this action is executed through Prolog, where $\widehat{q}$ is a Prolog query
   and $r$ is obtained through unification with the fact base representing the agent's state.
}.
The result
$v = f(s)(r)$, if it is not null, will be later checked if it solves the current goal
of the agent.

Figure \ref{example.dumb} shows an example of an action to put a random cross on
an empty square in a GoMoku board, if there is an empty square left. Indeed, this is not a very
intelligent move. But the thing to note here is the action's guard. It inspects
the environment to see if the game board still has empty squares (lines \ref{inspect.env1}-\ref{inspect.env2}).
If so, a random one, say $sq'$, will be returned (line \ref{return.square}). When
the action is executed, this $sq'$ will be passed to the function in the
${\bf do\_}$-part, bound to the $sq$ parameter. In this example, this will in turn
call $\sf move$, which will then place a cross on this square $sq'$.

\begin{figure}
\begin{lstlisting}[mathescape=true,
    numbers=right,
    numberstyle=\tiny,
    numbersep=0pt
    ]
var dumb = action("dumb").
 . do_((AgentState s) $\rightarrow$ (Square sq) $\rightarrow$ { $\label{dumb.behavior}$
      s.env.move(CROSS, sq.x, sq.y) ;
      return s.env.crossWin() }
   )
 . on_((AgentState s) $\rightarrow$ {        $\label{dumb.guard}$
      var empties = s.env.emptySquares() ;   $\label{inspect.env1}$
      if (empties.size()==0) return null ;   $\label{inspect.env2}$
      return empties.get(rnd.nextInt(empties.size())) } $\label{return.square}$
   ) ;
\end{lstlisting}
\caption{\em An action that would randomly put a cross in an empty square
in a GoMoku board. As a side note, notice that we again use lambda-expressions
(lines \ref{dumb.behavior} and \ref{dumb.guard})
to conveniently introduce functions without having to create a class.
} \label{example.dumb}
\end{figure}

\subsection{Agent's deliberation cycle}

\vspace{-3mm}
Algorithm \ref{alg.agentcycle} shows how an \aplib\ agent executes. It runs in
typical BDI's sense-reason-act cycles, also known as {\em deliberation cycles} \cite{MeyerAgentTech2008,AgentDesignPat2016,rao1992abstract}.
 As we will see, \aplib\ allows goals and tactics
to be hierarchically structured. This provides a simple but powerful
means for programmers to strategically program their agents, but on the other hand
an agent now has additional tasks, namely to keep track of its current and next goal and tactic
within the aforementioned hierarchies, as well as to regulate budget allocation.
Consequently, Algorithm \ref{alg.agentcycle} is more
elaborate than the base BDI algorithm as in \cite{rao1992abstract}.

Imagine an agent $A = (s_{{\mapsto} E},\Pi,\beta)$.
The execution of $A$ proceeds discretely in {\em ticks}. It sleeps between ticks (line \ref{sleep}),
though an incoming message will awaken it.

At the start, $A$ inspects its goal structure $\Pi$ to determine which goal $g_{{\mapsto} T}$ in $\Pi$ it should
pursue (line \ref{label.decidegoal1}). In the example in Fig. \ref{fig.createagent} $\Pi$
consists of only a single goal, so this is the one that will be selected. $A$ then
calculates how much of its budget $\beta$ should
be allocated for solving $g$ ($\beta_g$). $A$ will then pursue $g$. This means
{\em repeatedly} applying $T$ over multiple ticks until $g$ is solved, or
$\beta_g$ is exhausted. In BDI terminology, this reflects the promotion of $g$ from goal to
intent.

\begin{algorithm}
    \SetKwFor{ForEach}{on}{do}{}
    Let $A = (s_{{\mapsto} E},\Pi,\beta)$ be an agent.

    $g_{{\mapsto} T} \leftarrow {\sf obtainCurrentGoal}(\Pi)$ \label{label.decidegoal1}

    $\beta_g \leftarrow $ allocate budget for $g$ from $\beta$

    \While{$g \not= \bf null$}{ \label{agentloop}

         \If{$\beta_g > 0$} {   \label{checkbudget}
              $E.{\sf refresh}()$   // sensing the environment \label{sensing}

              {\color{blue}$actions \leftarrow {\sf obtainEnabledActions}(T,s)$}  \label{obtainEnabled}

              \If{$actions \not= \emptyset$} {
                   $\alpha \leftarrow {\sf choose}(actions)$  \label{chooseaction}

                   $v \leftarrow \alpha.{\sf execute}()$

                   \For{ {\bf each }$G \in g.{\sf ancestors}()$}{
                      {\color{blue} $\beta_G \leftarrow \beta_G$ - $\alpha$'s comp. cost } \label{substractingcost}
                   }


                   \If{$v \not= {\bf null} \wedge g.{\sf evaluate}(v) = {\bf true}$} { \label{check.goal}

                       mark $g$ as solved.

                       $g_{{\mapsto} T} \leftarrow {\sf obtainCurrentGoal}(\Pi)$

                       $\beta_g \leftarrow $ allocate budget for $g$ from $\beta_\Pi$

                   }
                   \Else{
                      {\color{blue} $T' \leftarrow {\sf next}(\alpha) \; ; \; T \leftarrow T'$} \label{get-next-subtactic}
                   }
              }
              \If{$g \not= \bf null$} {
                  sleep until a $tick$ or a message arrives. \label{sleep}
              }
          }
          \Else{
              mark $g$ as failed.

              $g_{{\mapsto} T} \leftarrow {\sf obtainCurrentGoal}(\Pi)$

              $\beta_g \leftarrow $ allocate budget for $g$ from $\beta$
          }
    }
\caption{\em The execution algorithm of an \aplib\ agent.
   Lines \ref{obtainEnabled} and \ref{get-next-subtactic} (blue)
   will be elaborated in Section \ref{sec.tactic}, and line \ref{substractingcost}
   in Section \ref{sec.structuredgoal}.
   } \label{alg.agentcycle}
\end{algorithm}

A single cycle of $A$'s execution is a single iteration of the loop in line \ref{agentloop}.
These are essentially what the agent does every cycle:

\begin{enumerate}

\item {\em Sensing.} The agent starts a cycle by
sensing the environment (line /ref{sensing}).
This updates $E$'s state, and hence also the agent's state $s$.

\item {\em Reasoning.}
To make itself responsive to changes in
the environment, an agent only executes one action per cycle. So, if
the environment's state changes at the next cycle, a different action
can be chosen to respond to the change.
Lines  \ref{obtainEnabled}-\ref{chooseaction} represent the agent's
reasoning to decide which action is the best to choose.

Let $g_{{\mapsto} T}$ be the agent's current goal, and $T$ is the tactic that is
associated with it to solve it. In the simple case, $T$ is just a single
action like in (\ref{ex.fluent.interf}), though generally it can be
composed from multiple actions. The agent determines which
actions in $T$ are enabled on the current state $s$ (line \ref{obtainEnabled}).
An action $\alpha$ is enabled on $s$ if it is eligible for
execution on that state. Roughly, this means that its guard yields a
non-null value when evaluated on $s$; we will refine this definition later.
If this results in at least one action, the method $\sf choose$ will choose
one. The default is to select randomly. If no action in $T$ is enabled,
the agent will sleep (line \ref{sleep}), hoping that at the next cycle the
environment state changes, hence enabling some actions.

\item {\em Execution and resolution.}
Let $\alpha$ be the selected action. It is then executed. If its result
$v$ is non-null, it is considered as a candidate solution to be checked against
the current goal $g$ (line \ref{check.goal}). If the goal
is solved, the agent inspects the remaining goals in $\Pi$ to decide
the next one to handle, and the whole process is repeated again with the new
goal. If there is no goal left, then the agent is done.

If $v$ does not solve the goal, at the next cycle the agent will again
select which action to execute (Line \ref{obtainEnabled}). It may
choose a different action. Things are now different than in
non-tactical BDI agents. 2APL or GOAL agents use a flat structured
plan-base, hence they always choose from the whole set of available plans/actions.
In \aplib, the structure of a tactic statically limits the choice of
eligible actions (while actions' guards dynamically refine the choice).
Deciding the next action to choose is therefore done in two stages: Line \ref{get-next-subtactic}
inspect the tactic tree to first select which enclosing subtactic $T'$ is eligible for the next cycle.
The agent then sleeps until the next tick (or until a message arrives).
Then, when the next cycle starts, Line \ref{obtainEnabled} gathers all guard-enabled candidate actions
within this $T'$, and then we do the rest of the cycle in the same way as before.

\end{enumerate}

\vspace{-4mm}
\section{Tactic}\label{sec.tactic}

\vspace{-3mm}
Rather than using a single action, \Aplib\ provides a more powerful means
to solve a goal, namely tactic. A tactic is a hierarchical composition of
actions. Methods used to compose them are also called {\em combinators}.
Figure \ref{fig.tactic} shows an example of composing a tactic, using $\sf FIRSTof$ and $\sf ANYof$
as combinators.
Structurally, a tactic is a tree with actions as leaves and tactic-combinators as nodes.
The actions are the ones that do the actual work. Furthermore, recall that the actions also have their own
guards, controlling their enabledness. The combinators are used to exert a higher level control over the actions,
e.g. sequencing them, or choosing between them. This higher level control supersedes guard-level control\footnote{While it is true that we can encode all control in action guards, this would not be an abstract way of programming tactical
control and would ultimately result in error prone code.}.

\begin{figure}
\begin{lstlisting}[mathescape=true,
    numbers=left,
    numberstyle=\tiny,
    numbersep=5pt
    ]
var T = FIRSTof(
    action("win1")  .do_(..).on_(winInOneMove).lift()
  , action("defend").do_(..).on_(danger).lift()
  , ANYof($\alpha_1$, $\alpha_2$)
  )
\end{lstlisting}
\caption{\em Defining a tactic $T$ for the GoMoku agent in Fig. \ref{fig.createagent},
composed of three other tactics. The combinator $\sf FIRSTof$ will choose
the first sub-tactic that is enabled for execution.
The tactic "win1" is an action (the code is not shown) that would do a winning move, if win is achievable in one move.
The tactic "defend" is also action; it will block the opponent if the latter can eminently move into a winning
configuration. If winning in one move is not possible, and the agent is not in eminent danger of losing,
the 3nd sub-tactic randomly chooses between two actions $\alpha_1$ and $\alpha_2$.
} \label{fig.tactic}
\end{figure}

The following tactic combinators are provided; let $T_1,...,T_n$ be tactics:

\begin{enumerate}

    \item If $\alpha$ is an action, $T = \alpha.{\bf lift}()$ is a tactic.
    Executing this tactic on an agent state $s$ means executing $\alpha$ on $s$. This is
    of course only possible if $\alpha$ is enabled on $s$
    (if its guard results a non-null value when queried on $s$).
    The execution of an action always takes a single tick.

    \item $T = {\bf SEQ}(T_1,...,T_n)$ is a tactic.
    When invoked, $T$ will execute
    the whole sequence $T_1$, ..., $T_n$. This will take at least $n$ ticks
    time (exactly $n$ ticks if all $T_i$'s have no deeper $\bf SEQ$ construct).

    \item $T = {\bf ANYof}(T_1,...,T_n)$ is a tactic that randomly chooses one of executable/enabled
    $T_i$'s and executes it. The 'enabledness' of tactics will be defined later.

   \HIDE{
    Notice that using ${\sf ANYof}$ does not mean the agent will just
    randomly select its sub-tactics. The random selection is {\em limited}
    over enabled sub-tactics.
    Since the enabledness of a tactic is ultimately determined by the guards of
    their actions, the choice would for the most part be driven by the agent's (its actions')
    own logic. Only on those states where the logic really does not know which actions/tactics
    to choose, then ${\sf ANYof}$ will make the decision.
    }

    \item $T = {\bf FIRSTof}(T_1,..,T_n)$ is a tactic. It is used to express priority
    over a set of tactics if more than one of them could be enabled.
    When invoked,
    $T$ will invoke the first enabled $T_i$ from the sequence $T_1,..,T_n$.

\end{enumerate}

Consider a goal $g_{{\mapsto} T}$. So, $T$ is the specified tactic to solve $g$. Recall that this means
that the agent will repeatedly try $T$, over possibly multiple ticks, until $g$ is solved or until
$g$'s budget runs out. So, the execution of a tactic is implicitly always iterative. If $T$
contains $\bf SEQ$ constructs, these will require the corresponding sub-tactics to be executed in
sequence, hence introducing inner control flows that potentially spans over multiple ticks as well.
This makes the execution flow of a tactic non-trivial. Let us therefore first introduce some support concepts.

\HIDE{
Because a tactic can sequence multiple actions, we say that its execution has {\em completed a cycle}
if all the actions that it is supposed to execute in sequence have been executed.
Executing an action takes a tick/deliberation cycle (not to be confused with tactic cycle meant above).
So, completing a single cycle of a tactic may take multiple
ticks. When $T$ is bound to a goal (using the
method $\sf tactic$ as in Figure \ref{fig.createagent}), the executing agent will repeatedly apply $T$
in potentially multiple cycles, until the goal is solved, or until the agent
decides to drop the goal, or it has exhausted the credit allocated for the corresponding goal.
}

If $T$ is a tactic and $s$ is the current agent state, ${\sf first}(T,s)$
is the set of actions in $T$ that are eligible as the first action to execute
to start $T$, and are furthermore enabled in $s$. $T$ is said to be {\em enabled} on
$s$ if ${\sf first}(T,s)\not=\emptyset$. Obviously, a tactic can only be invoked
if it is enabled. Since enabledess is defined in terms of $\sf first$, it is sufficient
to define the later:

\begin{mydef}
${\sf first}(T,s)$ is defined recursively as follows:

\begin{itemize}
\item ${\sf first}(\alpha.{\bf lift}(),s) = \{ \alpha \}$, if $\alpha$ is enabled on $s$, else
    it is $\emptyset$.
\item ${\sf first}({\bf SEQ}(T_1,...,T_n),s) = {\sf first}(T_1,s)$.
\item ${\sf first}({\bf ANYof}(T_1,...,T_n),s)$ is the union of ${\sf first}(U,s)$,
for all $U{\in}\{ T_1,...,T_n\}$.
\item ${\sf first}({\bf FIRSTof}(T_1,...,T_n),s)$ is ${\sf first}(T_1,s)$, if $T_1$ is enabled on $s$,
    else it is equal to ${\sf first}({\bf FIRSTof}(T_2,...,T_n),s)$, if $n\geq 2$, and else
    it is $\emptyset$.
\end{itemize}
\end{mydef}

Let $\alpha$ be an action in a tactic $T$. After $\alpha$ is completed, the agent will need to determine
which action to do next. This is not only determined by the enabledness of the actions, but also the
tactic sequencing imposed by $\bf SEQ$ and $\bf FIRSTof$ that are present in $T$. If $U$ is a sub-tactic,
let us define ${\sf next}(U)$ to be the next tactic that has to be executed after $T$ is completed.
Then in follows that the next action after $\alpha$ if ${\sf first}({\sf next}(\alpha),s)$, where $s$
is the agent's current state. The definition is below:

\begin{mydef}
Let $U$ be a tactic. Since a tactic syntactically forms a tree, every sub-tactic, except the root, in this tree has
a unique parent.
${\sf next}(U)$ is defined recursively as follows. Let $U' = {\sf parent}(U)$.

\begin{itemize}
\item If $U'$ is ${\bf SEQ}(T_1,...,T_n)$ and $U = T_i$, $i{<n}$, then ${\sf next}(U) = T_{i{+}1}$.
If $U = T_n$, then ${\sf next}(U) = {\sf next}(U')$.

\item If $U'$ is ${\bf ANYof}(T_1,...,T_n)$ then ${\sf next}(U) = {\sf next}(U')$.

\item If $U'$ is ${\bf FIRSTof}(T_1,...,T_n)$ then ${\sf next}(U) = {\sf next}(U')$.

\item If $U$ has no parent (so it is the root tactic), then ${\sf  next}(U) = U$.
\end{itemize}
\end{mydef}

Now we can define how the tactic in $g_{{\mapsto} T}$ is executed. When the goal
is first adopted the first actions eligible for execution are those from
${\sf first}(T,s)$ where $s$ is the agent current state.
In Algorithm \ref{alg.agentcycle} this is calculated in line \ref{obtainEnabled}.
The function ${\sf obtainEnabledActions}(T,s)$ is thus just ${\sf first}(T,s)$.

Suppose $\alpha{\in}{\sf first}(T,s)$ is selected. After this is executed,
the agent first calculate which sub-tactic of $T$ it should next execute.
This is calculated by $T' \leftarrow {\sf next}(\alpha)$ in line \ref{get-next-subtactic} in
Algorithm \ref{alg.agentcycle}. When the new cycle starts, the next set of
actions eligible for execution would be ${\sf first}(T',s)$, which is again
calculated by line \ref{obtainEnabled}. This goes on until the goal is solved.
Notice than when all sequential sub-tactics of a top-level tactic $T$ have been executed (which
would take multiple ticks to do), the last case in the definition of ${\sf next}$
will return $T$ itself as the next tactic to execute, essentially reseting the
execution of $T$ to start from its first action again.

\vspace{-5mm}
\section{Reasoning}\label{sec.reasoning}

\vspace{-2mm}
Most of agent reasoning is carried out by actions' guards, since they are the ones
that inspect the agent's state to decide which actions are executable.
Fig. \ref{example.dumb} showed an example of defining a simple action in \aplib.
Its guard (lines \ref{dumb.guard}-\ref{return.square}) queries the environment,
representing a GoMoku board, to obtain an empty square, if there is any.
The reader may notice that this query is imperatively formulated, which is to be
expected since \aplib's host language, Java, is an imperative programming language.
However, \aplib\ also has a Prolog backend
(using tuprolog \cite{2p-alpnews2013}) to facilitate a declarative style of state query.

Figure \ref{fig.prolog} shows an example. To use Prolog-style query, the agent's state
needs to extend the class $\sf StateWithProlog$. It will then inherit an instance of
a tuprolog engine to which we can add facts and inference rules, and
then pose queries over these. The example shows the definition of the action $\sf "win1"$
that we had in Fig. \ref{fig.tactic}, that is part of the tactic for the GoMoku agent
in Fig. \ref{fig.createagent}. The guard of this action searches for a move that would
win the game for the agent in a single step. This is formulated by the query in
line \ref{win1.guard}, which is interpreted as a Prolog-style query on the predicate
${\sf winningMove}(X,Y)$. This in turn is defined as a Prolog-style rule/clause in lines
\ref{winningMove.Start}-\ref{winningMove.End}. We do not show the full definition
of the rule, but for example lines \ref{horizontalStart}-\ref{horizontalEnd} characterize
four crosses in a row, and an empty square just left of the first cross, and hence
this empty square would be a solution for the predicate ${\sf winningMove}(X,Y)$
(putting a cross on this empty square would win the game for the agent).
Notice that the rule is declarative, as it only characterizes the
properties that a winning move/square needs to have;
it does not spell out how we should iterate over the game board in order to check it.

\begin{figure}
\begin{lstlisting}[mathescape=true,
    numbers=left,
    numberstyle=\tiny,
    numbersep=5pt
    ]
class AgentState extends StateWithProlog {   $\label{extend.StateProlog}$
 AgentState() {
   addRules(
     clause(winningMove("X","Y"))              $\label{winningMove.Start}$
     . IMPby(eastNeighbor(CROSS,"A","B","Y"))  $\label{horizontalStart}$
     . and(eastNeighbor(CROSS,"B","C","Y"))
     . and(eastNeighbor(CROSS,"C","D","Y"))
     . and(eastNeighbor(CROSS,"D","E","Y"))
     . and(not(occupied("A","Y")))
     . and("X is A")                           $\label{horizontalEnd}$
     . toString(),
       ... // the rest of winningMove's rules
     ) }                                   $\label{winningMove.End}$
}
var win1 = action("win1")
 . do_((AgentState s) $\rightarrow$
       (Result r) $\rightarrow$ {
          var x = intval(r.get("X")) ;
          var y = intval(r.get("X")) ;
          s.env.move(CROSS,x,y) ;
          return s.env.crossWin() })
 . on_((AgentState s) $\rightarrow$ s.query(st.winningMove("X","Y"))) $\label{win1.guard}$
\end{lstlisting}
\caption{\em The definition of the $"\sf win1"$ action in Fig. \ref{fig.tactic}. Its guard is
  formulated declaratively in the Prolog style.
} \label{fig.prolog}
\end{figure}

\vspace{-6mm}
\section{Structured Goal}\label{sec.structuredgoal}

\begin{figure}
    \begin{center}
    \includegraphics[scale=0.4]{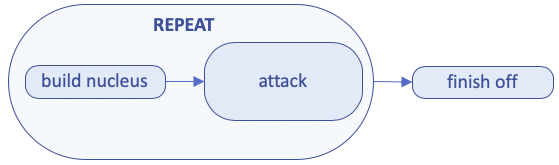}
    \end{center}
    \caption{\em A human strategy to win GoMoku. The player first tries to create a nucleus of
     enough number of his pieces. Then, he switches to attack
     to create a configuration where win is inevitable in at most
     two steps. If he manages to do this then it is a matter of finishing off the game.
     Else, if after sometime the attacking strategy cannot reach its goal,
     the player reset the strategy by trying to create a new nucleus.
    }\label{fig.strategy.gomoku}
\end{figure}

\vspace{-3mm}
A goal can be very hard for an agent to solve directly. It is then useful to
to provide additional direction for the agent e.g. in the form of subgoals.
For example, the GoMoku agent tactic in Fig. \ref{fig.tactic} is rather short sighted.
Its only winning strategy, $\sf win1$, is to detect a formation
where the agent would win in the next move and then to do this move.
An experienced opponent would prevent that the agent can create such a formation
in the first place. An example of a more sophisticated strategy is depicted
in Figure \ref{fig.strategy.gomoku}, involving repeating two stages until a formation
is created where win is inevitable no matter what the opponent does.
Tactics are not the right instrument to express such strategies. A tactic is intended to solve
a single goal, whereas the strategy in Figure \ref{fig.strategy.gomoku} consists of
multiple stages, each with its own goal.

In \aplib\ we can express such a strategy as a complex/composite goal
called a {\em goal structure}. It is a tree with goals as the leaves,
and goal-combinators as nodes.
The goals at the leaves are ordinary goals, and hence they all have tactics
associated to each. The combinators do not
have their own tactics. Instead, they are used to provide a high level
control on the order or importance of the underlying goals.

Available goal-combinators are as follows; let $G_1,...,G_n$ be goal structures:

\begin{itemize}
    \item If $g_{{\mapsto} T}$ is a goal with a tactic $T$ associated to it,
$g.{\sf lift}()$ will turn it to a goal structure consisting of
the goal as its only element.

    \item ${\bf SEQ}(G_1,...,G_n)$ is a goal structure that is solved by solving
all the subgoals $G_1,...,G_n$, and in that order.
This is useful when $G_n$ is hard to solve; so $G_1,...,G_{n{-}1}$ act as helpful intermediate goals
to guide the agent.

    \item $H = {\bf FIRSTof}(G_1,...,G_n)$ is a goal structure.
When given $H$ to solve, the agent will first try to solve $G_1$.
If this fails, it tries $G_2$, and so on until there is one goal $G_i$
that is solved. If none is solved, $H$ is considered as failed.

    \item If $G$ is a goal structure, so is $H = {\bf REPEAT}\;G$. When
    given $H$ to solve, the agent will pursue $G$. If after sometime
    $G$ fails, e.g. because it runs out of budget, it will be tried again.
    Fresh budget will be allocated for $G$, taken from what remains of
    the agent's total budget. This is iterated until $G$ is solved, or until
    $H$'s budget runs out.

\end{itemize}

We can now express the strategy in Fig. \ref{fig.strategy.gomoku} with
a goal structure of the form:
{\small
\[ {\bf var}
   \begin{array}[t]{l}
       gomGoal = \\
       \ \ \ {\bf SEQ}( \\
       \ \ \ \ \ \ {\bf REPEAT}({\bf SEQ}(
          \begin{array}[t]{l}
              G_1  \mbox{\ \ // create a nucleus}, \\
              G_2  \mbox{\ \ // attack} )),
          \end{array} \\
       \ \ \ \  \ \ G_3) \mbox{\ \ // finish off opponent}
  \end{array}
\]}

\subsubsection*{Dynamic Subgoals}

While there are plenty of problems that can be solved by decomposing it to 
a goal structure that remains unchanged through out the execution of
the agent, for greater strength and more flexibility \aplib\ agents can also
dynamically insert new sub-goal-structures into its goal structure.

Let $A$ be an agent and $H$ a goal structure.
The method $A.{\sf addAfter}(H)$ will insert $H$ as a next sibling of $A$'s current goal.
For example, if $\Pi = {\sf SEQ}(g_0,g_1)$ is $A$'s goal structure and $g_0$ is
the current goal, $A.{\sf addAfter}(H)$ will change $\Pi$ to ${\sf SEQ}(g_0,H,g_1)$.
This is useful when the agent, upon inspecting the current state of the environment, 
concludes that in order to later solve the next goal $g_1$ it is better
to first solve $H$, so it introduces $H$ as a new intermediate goal structure.

In a different situation $g_0$ fails and the agent $A$ notices that this is because
some necessary condition is not met. What it can do is to restart the attempt
to solve $g_0$, but this time inserting a new goal structure $H$ aimed at establishing
the missing condition. $A$ can do so by invoking $A.{\sf addBefore}(H)$.
Note that,= simply changing $\Pi$ to ${\sf SEQ}(H,g_0,g_1)$
will not work, because the behavior of ${\sf SEQ}$ dictates that the whole ${\sf SEQ}$
fails if one of its sub-goal-structure fails. So instead, ${\sf addBefore}(H)$ changes
$\Pi$ to ${\sf SEQ}({\sf REPEAT}({\sf SEQ}(H,g_0)),g_1)$.
The ${\sf REPEAT}$ construct will cause the agent to move back to $H$ upon faiing $g_0$.
The sequence ${\sf SEQ}(H,g_0)$ will then be repeatedly attempted until it succeeds.
The number of attempts can be controlled by assigning budget to the 
${\sf REPEAT}$ construct (budgeting will be discussed below).

\subsubsection*{Budgeting}

Since a goal structure can introduce multiple goals, they will be competing for
the agent's attention. By default, \aplib\ agents use the blind commitment
policy \cite{BDICh2015} where an agent will commit to its current goal
until it is solved. However, it is possible to exert finer control on
the agent's commitment through a simple but powerful budgeting mechanism.


Let $\Pi$ be the root goal structure that is given to an agent to solve.
For each sub-structure $G$ in $\Pi$ we can specify
a maximum on the budget it will get. Let us denote this by $G.{\sf bmax}$.
If left unspecified, the agent conservatively assumes that $G.{\sf bmax} = \infty$.
By specifying $\sf bmax$ we control how much the agent should commit to a particular
goal structure. This simple mechanism allows budget/commitment to be specified at the goal level (the leaves
of $\Pi$), if the programmer really wants to micro-manage the agent's commitment, or higher
in the hierarchy in $\Pi$ if he prefers to strategically control it.

When the agent was created, we can give it a certain initial computing budget $\beta_0$. If this is
unspecified, it is assumed to be $\infty$. Once it runs, the agent
will only work on a single goal (a leaf in $\Pi$) at a time.
The goal $g$ it works on is called the {\em current goal}. This also implies that
every ancestor goal structure $G$ of $g$ is also current. For every goal structure
$G$, let $\beta_G$ denote the remaining budget for $G$. At the beginning,
$\beta_\Pi = \beta_0$.

When a goal  or goal structure $G$ in $\Pi$ that was not current becomes current,
budget is allocated to it as follows, When $G$ becomes current, its parent either
becomes current as well, or it is already current (e.g. the root goal structure $\Pi$
is always current).
Ancestors $H$ that do not become current because they are already current will keep their
budget ($\beta_H$ does not change).
Then, budget for $G$ is allocated by setting $\beta_G$ to
${\bf min}(G.{\sf bmax},\beta_{{\sf parent}(G)})$, after we recursively determine $\beta_{{\sf parent(}G)}$.
Note that this budgeting scheme is {\em safe} in the sense that the budget of a goal structure
never exceeds that of its parent.

When working on a goal $g$, any work the agent does will consume some budget, say $\delta$. This will
be deducted from $\beta_g$, and likewise from the budget of other goal structures which are current
(line \ref{substractingcost} in Algorithm  \ref{alg.agentcycle}).
If $\beta_g$ becomes 0 or negative, the agent aborts $g$ (it is considered as failed).
It will then have to find another goal from $\Pi$.
Since the budget of a goal structure is at most equal to that of its parent, the lowest level
goal structure (so, a goal such as $g$ above) is always the first that exhausts its budget.
This justifies line \ref{checkbudget}
in Algorithm  \ref{alg.agentcycle} that only checks the budget of the current goal.

Depending on the used budgeting unit it may or may not be possible to guarantee
that $\beta_G$ will never be negative. If this can be guaranteed, the above
budgeting scheme also guarantees that the total used budget will never
exceed ${\bf min}(\Pi.{\sf bmax},\beta_0)$.

\vspace{-5mm}
\section{Related Work} \label{sec.relatedwork}

\vspace{-3mm}
To program agents, without having to do everything from scratch, we can either use
an agent 'framework', which essentially provides a library, or we use a dedicated
agent programming language. Examples of agent frameworks
are
JADE \cite{bellifemine1999jade} for Java,
HLogo \cite{bezirgiannis2016hlogo} for Haskell,
and PROFETA \cite{fichera2017python} for Python. Examples of dedicated agent
languages are
   JASON \cite{bordini2007programming}, 
   2APL \cite{dastani20082apl},
   GOAL \cite{GOALprogmanual2018},
   JADEL \cite{iotti2018agent},
   and SARL \cite{iat2014}.
HLogo is an agent framework that is more specialized for developing an agent-based
simulation, which means that HLogo agents always operate on a fixed albeit configurable environment,
namely the simulation world.
On the other hand, JADE is a generic agent framework that can be connected to any environment.
\Aplib\ is also a generic agent framework, however it has been designed to offer the
fluency of an embedded Domain Specific Language (DSL).
It makes heavy use of design patterns such as Fluent Interface \cite{fowler2005fluent}
and Strategy Pattern \cite{designpatterns94} to improve its fluency.
\Aplib\ is light weight compared to JADE. E.g. the latter supports distributed
agents and FIPA compliance\footnote{
FIPA (\url{http://www.fipa.org/}) defines a set of standards for interoperation of heterogeneous agents.
While the standards are still available, FIPA itself is no longer active as an organization.
} which \aplib\ do not have.
JADE does not natively offers BDI agency,
though BDI agency, e.g. as offered by 2APL and JADEL, can be implemented on top of JADE.
In contrast, \aplib, and PROFETA too, are natively BDI agent frameworks.

Among the dedicated agent programming languages, JASON, 2APL, and GOAL are dedicated for programming
BDI agents. In addition to offering BDI concepts such as beliefs and goals, these languages
also offer Prolog-style declarative programming. They are however rather restricted in available
data types (e.g. no support for collection and polymorphism). This is a serious hinderance if
we are to use them for large projects. JADEL and SARL are non-BDI. In particular SARL has
a very rich set of language features (collection, polymorphism, OO, lambda expression).
PROFETA, and \aplib\ too, are somewhere in between. Both are BDI DSLs, but they are
embedded DSLs rather than a native language as SARL. Their host languages are full of
features (Python and Java, respectively), that would give the strength of SARL that
agent languages like JASON and GOAL cannot offer.

\Aplib's distinguishing feature compared to other implementations of BDI agency (e.g. JACK, JASON, 2APL, GOAL, JADEL,
PROFETA) is its tactical programming of plans (through tactics) and goals (through goal structures).
An agent is essentially set of actions. The BDI architecture does not traditionally impose a rigid control structure
on these actions, hence allowing agents to react adaptively to changing environment.
However, there are also goals that require certain actions to be carried out in
a certain order over multiple deliberation cycles. Or, when given a hard goal to
solve, the agent might need to try different strategies, each would need to be given
enough commitment by the agent, and conversely it should be possible to abort it
so that another strategy can be tried. All these imply that tactics and strategies require
some form of control structures, although not as rigid as in e.g. procedures. All
the afore mentioned BDI implementations do not provide control structures beyond
intra-action control structures.
This shortcoming was already observed by
\cite{evertsz2015framework}, stating domains like autonomous vehicles need
agents with tactical ability. They went even further, stating
that Agent Oriented Software Engineering (AOSE)
methodologies in general
do not provide a sufficiently rich representation of goal control structures.
While inter-actions and inter-goals control structures
can be encoded through pushing and popping of beliefs or goals into the agent's state,
such an approach would clutter the programs and error prone.
An existing solution for tactical programming for agents is to use the Tactics Development
extension \cite{evertsz2015framework} of the Prometheus agent development methodology \cite{padgham2005prometheus}.
This extension allows tactics to be graphically modelled, and template implementations
in JACK can be generated from the models.
In contrast, \Aplib\ provides the features directly at the programming level.
It provides the additional control structures suitable for
tactical programming over the usual rule-based style programming of BDI agents.

We also want to mention FAtiMA \cite{dias2014fatima}, which is a BDI agent framework, but it extends
agents' BDI state with emotional states. At the first glance, emotion and tactical
thinking would be considered as complementary, in situations where an agent has to
work together with a human operator it would be reasonable to envisage the agent to take
the human's emotional state into account in its (the agent's) tactical decision making.
This can be done e.g. by deploying a FAtiMA agent whose task is to model
the user's emotional state. While interesting, such a combination requires further
research, and hence it is future work for us.

\vspace{-6mm}
\section{Conclusion \& Future Work} \label{sec.concl}

\vspace{-3mm}
We have presented \aplib, a BDI agent programming framework featuring multi agency and novel
tactical programming and strategic goal-level programming.
We choose to offer \aplib\
as a Domain Specific Language (DSL) embedded in Java, hence making the framework very expressive.
Despite the decreased fluency, we believe this embedded DSL approach to be better suited for
large scale programming of agents, while avoiding the high expense and long term risk of maintaining
a dedicated agent programming language.

While in many cases reasoning type of intelligence would work well, there are also cases where
this is not sufficient. Recently we have seen rapid advances in learning type of AI. As future work
we seek to extend \aplib\ to let programmers hook learning algorithms to their agents. This will allow them to
teach the agents to make the right choices, at least in some situations, which also means that
they can then program the agents more abstractly.

\bibliographystyle{apalike}
{\small
\bibliography{aplibBib}
}

\end{document}